\documentclass[paper]{ieice}
\usepackage[pdftex]{graphicx,xcolor}
\usepackage[fleqn]{amsmath}
\usepackage{newtxtext}
\usepackage[varg]{newtxmath}
\usepackage{multirow}

\field{}
\title{Multi-Scale Estimation for Omni-Directional Saliency Maps Using Learnable Equator Bias}
\authorlist{%
 \authorentry[takao-y@sophia.ac.jp]{Takao Yamanaka}{Member}{sophia}\MembershipNumber{0909050}
 \authorentry{Tatsuya Suzuki}{Nonmember}{sophia}\MembershipNumber{}
 \authorentry{Taiki Nobutsune}{Nonmember}{sophia}\MembershipNumber{}
 \authorentry{Chenjunlin Wu}{Nonmember}{sophia}\MembershipNumber{}
}
\affiliate[sophia]{The authors are with the Department of Information and Communication Sciences, Sophia University, Tokyo, 102-0094, Japan.}
\received{2023}{3}{22}
\revised{2023}{6}{10}

\definecolor{revised_color}{gray}{0}

\begin{document}
\maketitle
\begin{summary}
Omni-directional images have been used in wide range of applications including virtual/augmented realities, self-driving cars, robotics simulators, and surveillance systems. For these applications, it would be useful to estimate saliency maps representing probability distributions of gazing points with a head-mounted display, to detect important regions in the omni-directional images. This paper proposes a novel saliency-map estimation model for the omni-directional images by extracting overlapping 2-dimensional (2D) plane images from omni-directional images at various directions and angles of view. While 2D saliency maps tend to have high probability at the center of images (center bias), the high-probability region appears at horizontal directions in omni-directional saliency maps when a head-mounted display is used (equator bias). Therefore, the 2D saliency model with a center-bias layer was fine-tuned with an omni-directional dataset by replacing the center-bias layer to an equator-bias layer conditioned on the elevation angle for the extraction of the 2D plane image. The limited availability of omni-directional images in saliency datasets can be compensated by using the well-established 2D saliency model pretrained by a large number of training images with the ground truth of 2D saliency maps. In addition, this paper proposes a multi-scale estimation method by extracting 2D images in multiple angles of view to detect objects of various sizes with variable receptive fields. The saliency maps estimated from the multiple angles of view were integrated by using pixel-wise attention weights calculated in an integration layer for weighting the optimal scale to each object. The proposed method was evaluated using a publicly available dataset with evaluation metrics for omni-directional saliency maps. It was confirmed that the accuracy of the saliency maps was improved by the proposed method.
\end{summary}


\section{Introduction}\label{introduction}

Omni-directional images (ODI) are expected to be applied in widespread fields, such as virtual/augmented realities, self-driving cars, and robotics. It provides immersive impression in virtual environments using a head-mounted display (HMD). In addition, ODI can be also dealt with panorama viewers in personal computers or smart phones, which extend applicable situations of ODI\@. Estimation of salient regions in ODI will promote the applications of ODI \cite{ODIReview}. For example, it can be used for indicating detailed information of salient objects in virtual/augmented environments, or for suppressing amount of ODI data for network transmission. 

\begin{figure}[tb]
    \begin{center}
        \includegraphics[width=\linewidth]{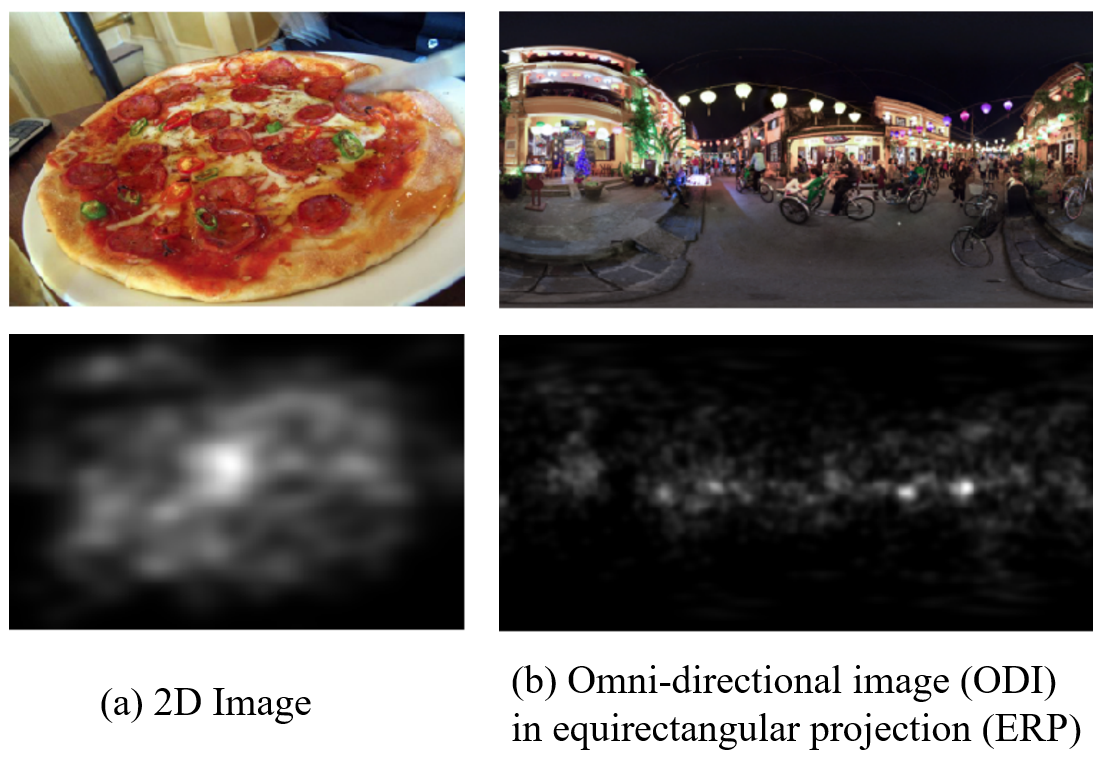}
    \end{center}
    \caption{Examples of saliency maps for 2D image and omni-directional image.}
    \label{fig:salmap_examples}
\end{figure}

ODI is usually represented in the equirectangular projection (ERP), as shown in Fig. \ref{fig:salmap_examples} (b). An approach to estimate ODI saliency maps is to directly apply a 2-dimensional (2D) saliency model to ODI in ERP \cite{abreu}. However, ODI in ERP has large distortion in the top and bottom areas (the north and south poles), which results in unsatisfactory saliency estimation. To overcome this distortion, the representation of cube mapping is also used for the saliency estimation in SalNet360 \cite{salnet360}, and then each cubic face is independently processed by a 2D saliency model, followed by a refinement stage to integrate saliency maps from all the faces. However, the cube mapping has too small field of view (90 degrees) to estimate accurate saliency in each cubic face \cite{qing2022}. In addition, an object is often divided into two or more cubic faces, so that it is difficult to detect the object by a 2D saliency model \cite{suzuki_SMC2018}.

To solve these problems, this paper proposes a novel ODI saliency model by extracting overlapping 2D plane images from ODI at various directions with multiple angles of view. 2D saliency models have been well established, and can be trained using large amount of training data \cite{mit,mit_tuebingen,densesal}, while ODI saliency datasets have been so far limited since obtaining fixation data is much more cumbersome than 2D images. By using the pretrained 2D saliency models, the required training data for the fine-tuning to the ODI dataset can be reduced. However, there is an important difference between the conventional 2D saliency model and the model for 2D plane images extracted from ODI\@. The saliency maps for normal 2D images have high prior probability at the center of images called center bias, while the high-probability region is observed at the horizontal directions in ODI saliency maps called equator bias \cite{sitzmann}, as shown in Fig. \ref{fig:salmap_examples}. Therefore, the 2D saliency model was pretrained with a learnable center-bias layer (single channel pixel-wise weighting layer) with 2D saliency datasets, and then the network was fine-tuned with an ODI dataset by replacing the center-bias layer to an equator-bias layer composed of pixel-wise weights in multiple channels, each of which corresponds to the elevation angle for the extraction of the 2D plane image. Furthermore, the proposed method extracts 2D plane images from ODI in multiple angles of view to detect objects of various sizes. Since the field of view changes depending on the angles of view, the receptive field to detect saliency for an object also changes depending on the angles of view. That is, information from larger area (receptive field) is used for larger angle of view. This would contribute on more accurate saliency detection in ODI\@. 

The contributions of this paper are as follows.
\begin{enumerate}
    \item A novel method to estimate the saliency maps for ODI was proposed by extracting overlapping 2D plane images from ODI at various directions with multiple angles of view, to accurately detect objects of various sizes with variable receptive fields.
    \item The ODI saliency model was realized based on a well-established 2D saliency model pretrained with large amount of training data, by learning prior distribution conditioned on elevation angles in ODI\@.
    \item It was confirmed from experiments that the accuracy of ODI saliency estimation was improved by the proposed method in a publicly available ODI saliency dataset. 
\end{enumerate}

\section{Related Works}\label{related_works}

\subsection{Saliency map estimation for 2D images}
\label{sec2:2dSaliency}
Many saliency models have been proposed to estimate the locations in 2D images which attract attentions of people. The saliency map computed from image features with these models represents the probability density function of fixations when people look at an image. In the earlier studies, the saliency map has been estimated by extracting low-level features. One of these saliency-map estimation models is that proposed by Itti et al.\ \cite{itti}, which estimates saliency maps by extracting early visual features such as intensities, colors, and orientations. AWS (Adaptive Whitening Saliency) \cite{aws} is another conventional model based on the whitening of low level features.

The recent progress of deep learning has also contributed to the improvement of saliency-map estimation models. Although the conventional models described above have used the hand-crafted features to estimate the saliency maps, the image feature vectors extracted from convolutional neural networks (CNN) have been used instead \cite{saliconnet,salnet,eDN,deepgaze1,deepfix,deepgaze2}. The ensembles of deep networks (eDN) \cite{eDN} have trained small-size CNN (up to 3 layers) to extract multiple-layer features for estimating the saliency maps by combining the features with support vector machines. DeepGaze I \cite{deepgaze1} has shown that the image features extracted from the multiple layers of CNN trained for an image-classification task on ImageNet \cite{imagenet} are useful to estimate the saliency maps. This implies that the transfer learning from the image classification to the saliency map estimation is effective because people tend to look at the centers of objects \cite{borji2016}, which are learned to recognize in the pretrained model for the ImageNet classification task. 

In contrast to DeepGaze I which uses AlexNet as the backbone network, SaliconNet \cite{saliconnet} and DeepFix \cite{deepfix} uses fully convolutional neural networks based on the VGG architecture which has shown better performance than AlexNet on the image classification task. SaliconNet \cite{saliconnet} estimates saliency maps with two image scales to allow the model to be robust against the sizes of objects in images. DeepFix \cite{deepfix} uses dilated convolution filters \cite{yu2016} to enlarge the receptive field and inception modules to capture multi-scale information. The deep-type network of SalNet \cite{salnet} has also adopted the VGG architecture with simplification composed of 10 layers. DeepGaze II \cite{deepgaze2} is also based on the VGG architecture, which uses a center-bias layer to incorporate the prior distribution and log-likelihood as the loss function. 
%Note that the VGG features of DeepGaze II are not fine-tuned with fixation datasets. These models have improved the performance over DeepGaze I in the MIT Saliency Benchmark \cite{mit}. 

In addition to the VGG-based CNN models, several CNN architectures have been used for the saliency-map estimation. For example, SalGAN \cite{salgan} is a model that estimates saliency maps using generative adversarial networks. DenseSal and DPNSal \cite{densesal} have shown better performance based on densely connected neural networks (DenseNet) \cite{densenet} and dual path networks (DPN) \cite{dpn}, respectively. EML-NET \cite{emlnet} combines the feature vectors extracted from DenseNet and NasNet \cite{zoph2018}. 
\textcolor{revised_color}{DeepGazeIIE \cite{deepgaze2e} also combines saliency predictions from multiple saliency models.}
These methods have achieved the state-of-the-art results on the MIT Saliency Benchmark \cite{mit} \textcolor{revised_color}{and MIT/Tuebingen Saliency Benchmark \cite{mit_tuebingen}}.

\subsection{Saliency map estimation for omni-directional images}
\label{sec2:ODISaliency}

In addition to the saliency-map estimation for 2D images, the saliency models for ODI have been developed since the ICME2017 Competition \cite{salient360_toolbox}. ODI is represented in various projections, among which ERP is commonly used to represent ODI in 2 dimensions as shown in Fig. \ref{fig:salmap_examples}, though it has distortions at poles (the top and bottom regions in ERP). The model by Abreu et al.\ \cite{abreu} has used SaliconNet \cite{saliconnet} to estimate the ODI saliency map by directly inputting ODI in ERP\@. The ERP saliency maps for horizontally different viewing directions are fused into an ERP saliency map to suppress the center-bias effect induced by the 2D saliency-map model (SaliconNet). However, the ERP images have distortions at poles, so that saliency at the poles cannot be correctly estimated. Moreover, the center-bias effect cannot be completely suppressed at the edges of the ERP image. 

Since the database for ODI saliency maps are relatively small, it would be important to first pre-train the CNN model for 2D images with large databases, and then fine-tune it with the ODI database. In SalNet360 \cite{salnet360}, this has been realized by subdividing ODI into 6 undistorted plane patches in cubic projection, estimating the 2D saliency maps with SaliconNet (2D saliency model) \cite{saliconnet} and a refinement network, and then integrating the 2D saliency maps into an ODI saliency map in ERP\@. In order to incorporate the dependence of saliency on the locations in ODI, the ODI spherical coordinates of each pixel in each 2D patch were input to the refinement neural network. However, since SaliconNet outputs a saliency map biased in the center of an image, the saliency at the 2D image edges is estimated lower. SalGAN360 \cite{salgan360} based on SalGAN \cite{salgan} has estimated the ODI saliency maps by fusing the saliency map in ERP and saliency maps for undistorted plane patches in the cubic projection to integrate the global and local information. This method employs multiple cubic projection, where overlapping cubic faces are extracted as in our proposed method. However, the angles of view are limited to 90$^\circ$ in the multiple cubic projection, and biases to represent prior distributions are not considered. Qing et al.\ \cite{qing2022} also have proposed a method to train a model with multiple cubic projection and to estimate saliency maps using ERP with multiple sphere rotation to reduce the influence of distortion at poles. Although global information can be utilized from ERP in the inference, the angle of view in the training stage is limited to 90$^\circ$ and the equator bias is not explicitly considered.

%% video saliency prediction for ODI
%The saliency maps has been also estimated for ODI videos. SaltiNet proposed by Assens et al.\ \cite{saltinet} is based on a temporal-aware novel representation of saliency information named the saliency volume, which is composed of feature maps with the temporal axis. However, this model does not consider the prior distribution and the distortions at the poles. Cheng et al.\ \cite{cubepadding} has proposed the model using the 6 undistorted patches with the cube padding, which uses the pixels of adjacent patches for padding in each layer of CNN to reduce the negative effects of dividing ODI\@.

%In this paper, a novel saliency-map estimation model for ODI is proposed by dividing ODI into the overlapping undistorted patches with multiple angles of view and modeling the prior distribution with a pixel-wise scaling layer depending on the elevation angle in the spherical coordinates.

\section{Method}\label{method}

\begin{figure}[tb]
    \begin{center}
        \includegraphics[width=\linewidth]{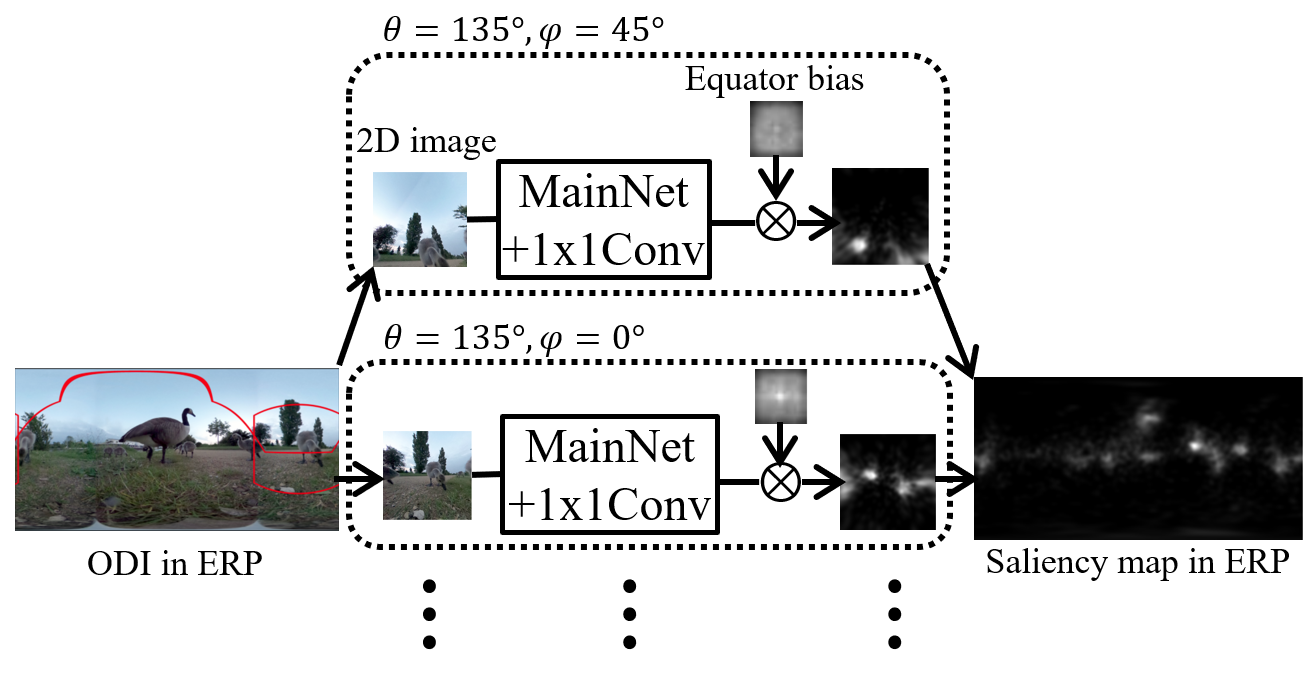}
    \end{center}
    \caption{ODI saliency estimation model by learning equator bias for each elevation angle. Network weights are shared among elevation angles except for equator bias.}
    \label{fig:odi_salmodel}
\end{figure}

\begin{figure}[tb]
    \begin{center}
        \includegraphics[width=\linewidth]{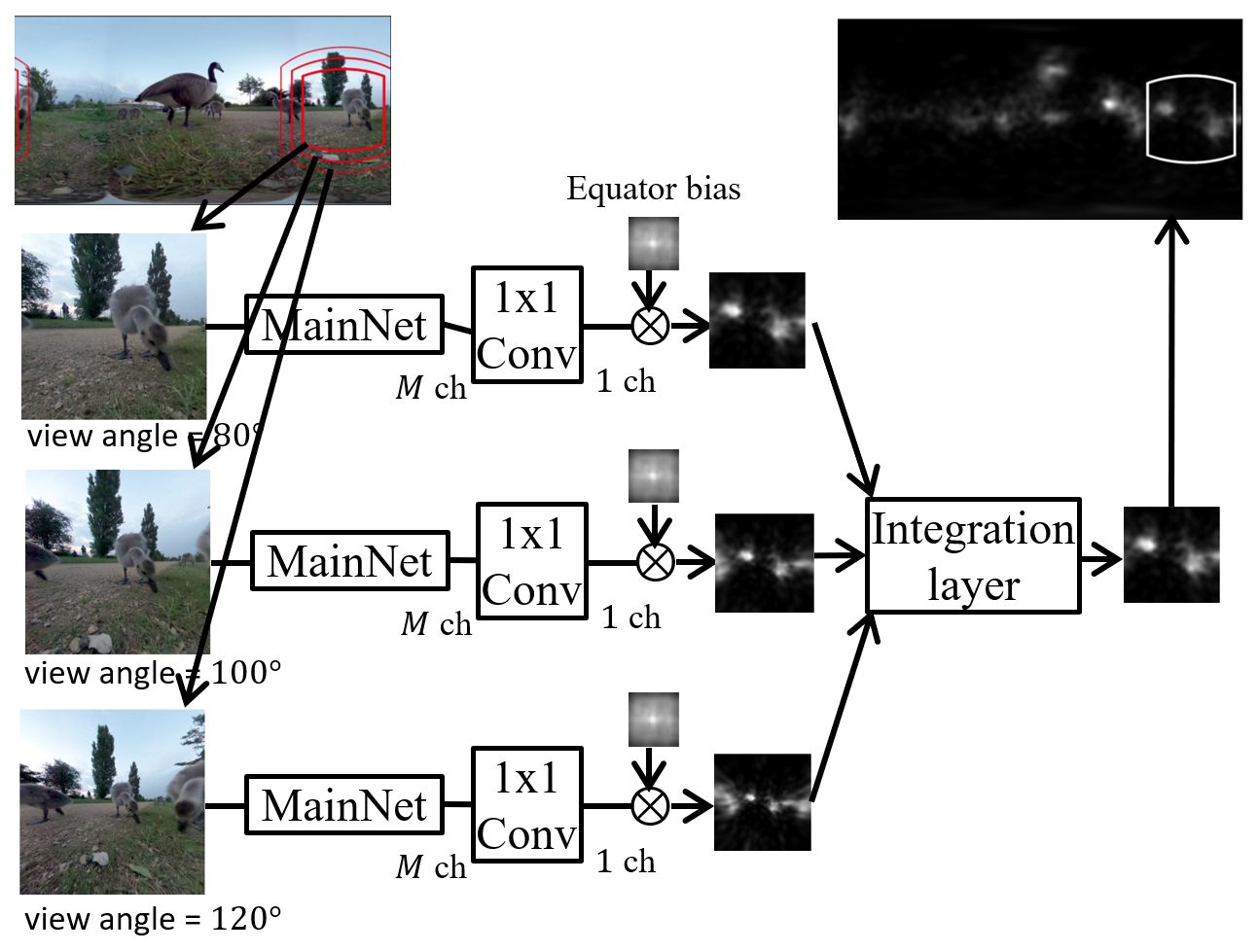}
    \end{center}
    \caption{Multi-scale saliency estimation model by extracting 2D images in multi-angles of view (\textcolor{revised_color}{dotted}-line block in Fig. \ref{fig:odi_salmodel}). Network weights are shared among multiple angles of view including equator bias.}
    \label{fig:mutiscale_salmodel}
\end{figure}

The proposed model for estimating ODI saliency maps is illustrated in Fig. \ref{fig:odi_salmodel} and Fig. \ref{fig:mutiscale_salmodel}. In the method, undistorted plane patches (2D images) are extracted in various viewing directions from ODI to use an existing saliency model for 2D images. These patches are overlapping each other in contrast to the cubic projection as in \cite{salnet360}. In addition, 2D patches are extracted in multiple angles of view for a viewing direction as shown in Fig. \ref{fig:mutiscale_salmodel}. Since an object appears in different sizes in these patches, the 2D saliency model can detect the object in the optimal scale with appropriate receptive field (information from surrounding region). Then, the patches in various viewing directions with multiple angles of view are fed into the 2D saliency-map estimation model (MainNet), to predict the saliency map for each patch. The 2D saliency-map estimation model is pretrained with large databases of 2D saliency maps such as the SALICON dataset \cite{salnet}. The outputs of MainNet are applied to the bias layer to model the prior distribution for ODI (equator bias) depending on the elevation angle in spherical coordinates. The patches in multiple angles of view are first integrated into a saliency map for a viewing direction by weighting important scales at each pixel using an integration layer consisting of an attention module, as in Fig. \ref{fig:mutiscale_salmodel}. Then, the integrated patches are fused into a saliency map in ERP, as in Fig. \ref{fig:odi_salmodel}. In the following subsections, key components in the proposed method are explained in detail.

%\subsection{ODI Saliency Model with Learnable Equator Bias}
\subsection{Bias Layer for Prior Distribution}
\label{sec3:BiasLayer}

\begin{figure}[tb]
    \begin{center}
        \includegraphics[width=\linewidth]{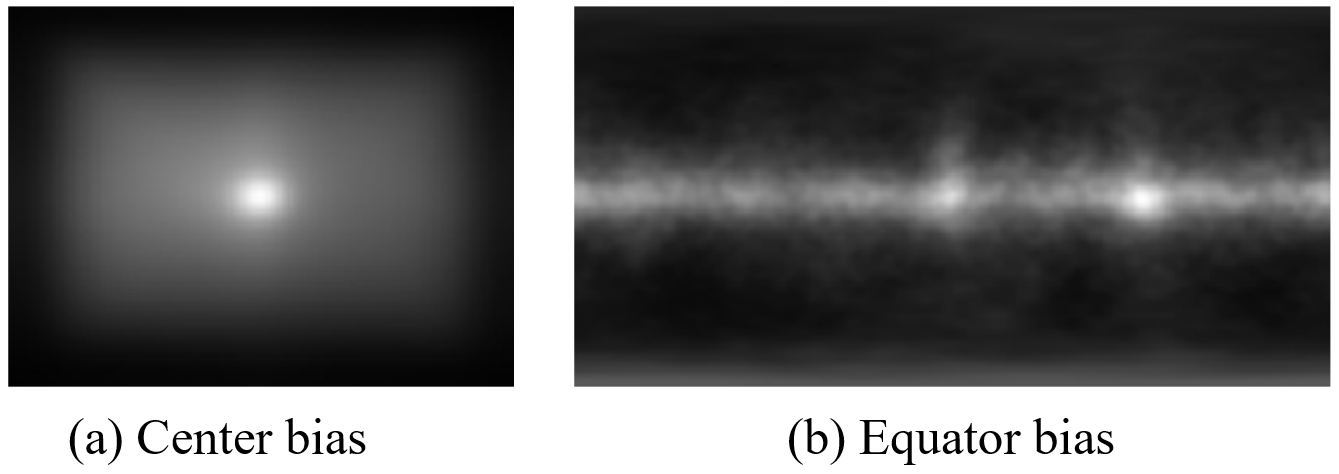}
    \end{center}
    \caption{Examples of biases: (a) center bias (average over 2D saliency maps in Salicon Dataset), (b) equator bias (average in Salient360 Dataset).}
    \label{fig:biases}
\end{figure}

As shown in Fig. \ref{fig:salmap_examples}, the center bias represents the property of tendency for fixations to concentrate on the center of a 2D image while the equator bias represents that to concentrate on the equator of ODI\@. The average of saliency maps over 2D images and ODI in databases are shown in Fig. \ref{fig:biases} (a) and (b), respectively. They represent the prior distributions of fixations, independent on the local image features. Thus, the prior distributions are different between 2D images and ODI, so that these differences have to be compensated in the ODI saliency-map estimation.

\begin{figure}[tb]
    \begin{center}
        \includegraphics[width=\linewidth]{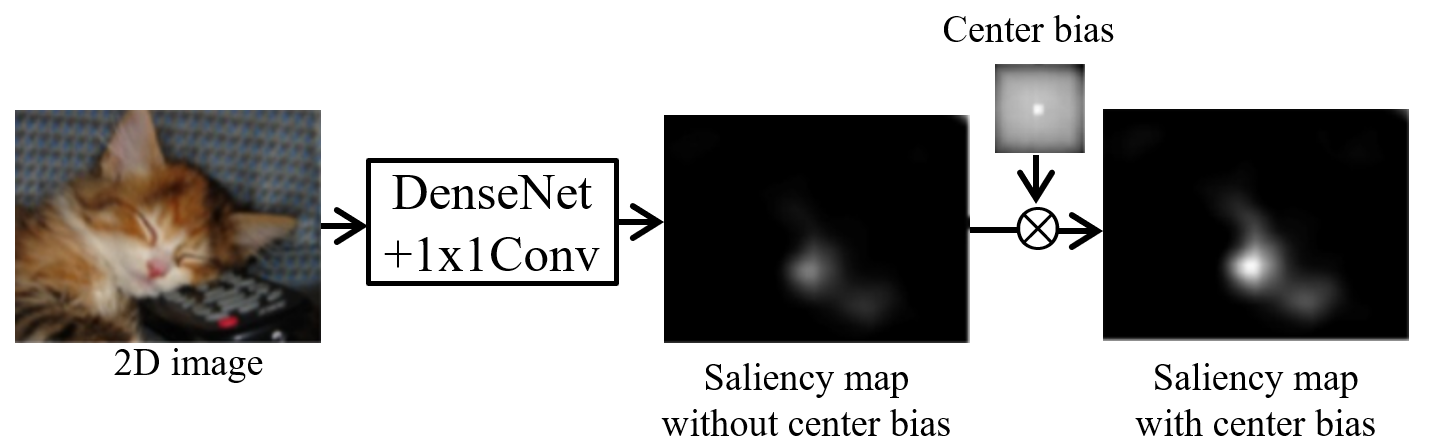}
    \end{center}
    \caption{2D saliency estimation model using center-bias layer.}
    \label{fig:2Dsalmodel}
\end{figure}

The saliency is defined as a probability of observing a fixation ($o$: binary random variable) at a location $x$ for a local image feature $f$: $p(o|f,x)$. If $x$ and $f$ are assumed to be independent, 
\begin{equation}
\begin{split}
    p(o|f,x) &= p(f|o,x)p(o|x)/p(f|x) \\
    &= p(f|o)p(o|x)/p(f) \\
    &= p(o|f)p(o|x)/p(o),
\end{split}
\end{equation}
where $p(o|f)$ represents the saliency map depending only on the local image feature $f$ without the prior probability $p(o|x)$, and $p(o)$ is a normalization constant. Thus, the saliency map can be modeled by the multiplication of $p(o|f)$ and $p(o|x)$ with normalization. In the proposed ODI saliency model, a 2D saliency-map estimation model, DenseSal \cite{densesal}, was used with an additional layer (bias layer) for learning the prior probability $p(o|x)$ in 2D saliency maps, as shown in Fig. \ref{fig:2Dsalmodel}. This layer consists of pixel-wise scaling weights for a feature map from the 2D saliency model. By learning the weights with 2D saliency datasets, the saliency depending only on a location $x$ in an image can be estimated in the center-bias layer. When estimating ODI saliency maps, this center-bias layer is replaced by an equator-bias layer, which consists of pixel-wise scaling weights in multiple channels corresponding to elevation angles of viewing directions for 2D patches.

\textcolor{revised_color}{
    In the 2D saliency model, the saliency at image boundaries are underestimated by the center-bias effect even if there are some objects at the edges of an image \cite{cornia2016,saldet2015}. In the extracted 2D images from ODI, objects can appear at the edge of the extracted images, which may be focused by people with HMD. Therefore, the center bias can give the negative effect to the saliency estimation for ODI. In our approach, first the 2D saliency model was trained with center-bias layer to model the center-bias effect existing in the 2D saliency dataset similar to \cite{cornia2016}, so that the MainNet (excluding the center-bias layer) represents only the saliency for contents (such as objects) not depending on the position of the image. By replacing the center-bias layer to the equator-bias layer, and by learning the bias layer using the training data extracted from ODI, the bias layer expresses the equator-bias effect (saliency depending on the elevation angle in the spherical coordinates, and not depending on the horizontal angle), so that the negative effect of the center bias can be eliminated. 
}

\subsection{Extraction of 2D Images from ODI}
\label{sec3:2DImageExtraction}

\begin{figure}[tb]
    \begin{center}
        \includegraphics[width=\linewidth]{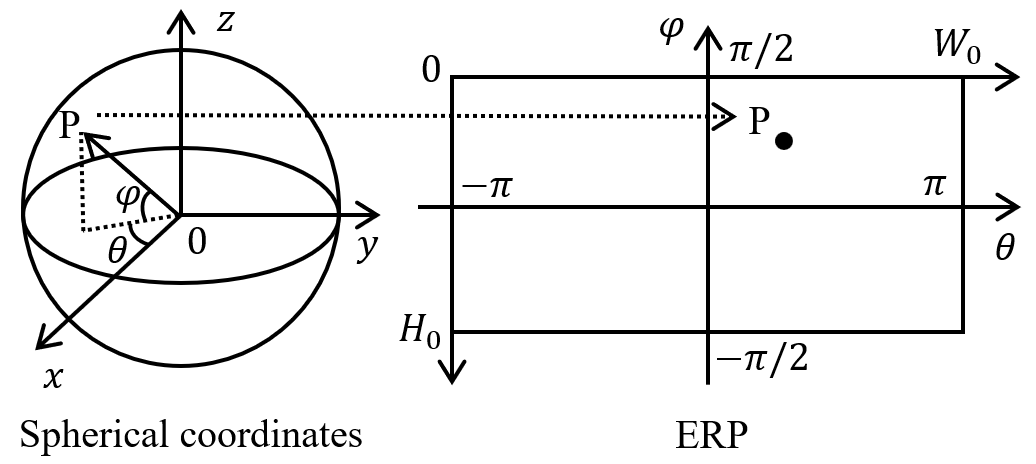}
    \end{center}
    \caption{Coordinate systems for ODI\@.}
    \label{fig:ODI_coord}
\end{figure}

Since ODI in ERP has distortion at poles as shown in Fig. \ref{fig:salmap_examples}, 2D images are extracted from ODI to estimate an ODI saliency map by removing the distortion. The correspondence between ERP and the spherical coordinate system is shown in Fig. 
\ref{fig:ODI_coord}. The unit vectors of a 2D image extracted at a viewing direction ($\theta_c, \phi_c$) in spherical coordinates are represented in the following equations in the 3D Euclidean coordinate system:
\begin{equation}
\begin{split}
    X_n &= (- \sin \theta_c, - \cos \theta_c, 0) \\
    Y_n &= (- \sin \phi_c \cos \theta_c, \sin \phi_c \sin \theta_c, \cos \phi_c)
\end{split}
\end{equation}
The ERP coordinates of the 2D image in a tangent plane can be obtained by transforming the 3D Euclidean coordinates of points in the 2D image to spherical coordinates. Thus, 2D images can be extracted from ODI in ERP using the coordinates. When 2D saliency maps are integrated into an ODI saliency map in ERP, each point in ERP is assigned to the nearest point in 2D saliency maps of multiple viewing directions.

\begin{figure}[tb]
    \begin{center}
        \includegraphics[width=\linewidth]{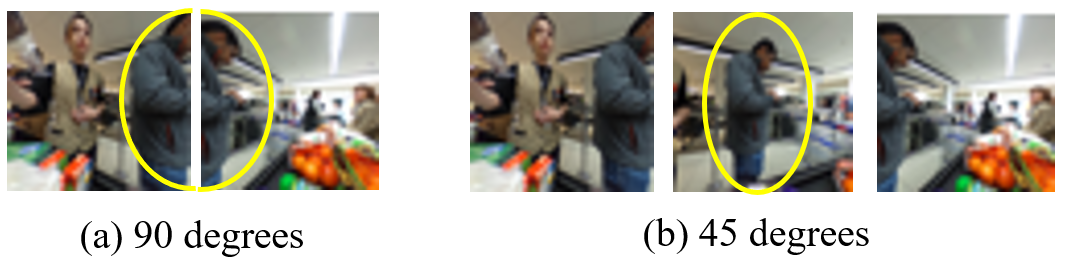}
    \end{center}
    \caption{Example images of different intervals of viewing directions. A human, who is cut off at the edge of the image in the viewing-direction interval of 90 degrees (a), can be detected in the interval of 45 degrees (b).}
    \label{fig:vd_interval}
\end{figure}

When the 2D images are extracted from ODI without overlapping (for example, cube mapping), the objects placed at the edge of the 2D patch are cut off as shown in Fig. \ref{fig:vd_interval} (a), so that it would be difficult to be recognized for accurate saliency estimation. However, if the 2D images are extracted from ODI with overlapping as in Fig. \ref{fig:vd_interval} (b), the objects can be detected, leading to more accurate estimation. In the integration of 2D saliency maps for multiple viewing directions, the saliency values in ERP are calculated by averaging the overlapping saliency values. Note that the 2D saliency maps with multiple angles of view are integrated using an integration layer by weighting the optimal scales, as explained in the following subsection and shown in Fig. \ref{fig:mutiscale_salmodel}.
\textcolor{revised_color}{
Although the multiple viewing directions can also be integrated using importance weights depending on the position of an extracted image (saliency at the center of an extracted image may be more accurately predicted than at the edge of an extracted image), it would be less effective than weighting optimal scales, so that a simple averaging integration was selected in the proposed method for the simplicity. 
}

\subsection{Multi-scale Saliency Model with Multiple Angles of View}

In the proposed method, 2D images are extracted in multiple angles of view, as shown in Fig. \ref{fig:mutiscale_salmodel}. The relationship between the angle of view ($\theta_a, \phi_a$) and the extracted 2D image size ($W, H$) can be represented by the following equation, where $L$ is the distance from the camera to the image plane.
\begin{equation} \label{eq:view_angle}
    \begin{split}
    \tan (\theta_a/2) &= W/2L \\
    \tan (\phi_a/2) &= H/2L
\end{split}
\end{equation}
The 2D image size ($W, H$) are set to constant values for multiple angles of view.

\begin{figure}[tb]
    \begin{center}
        \includegraphics[width=\linewidth]{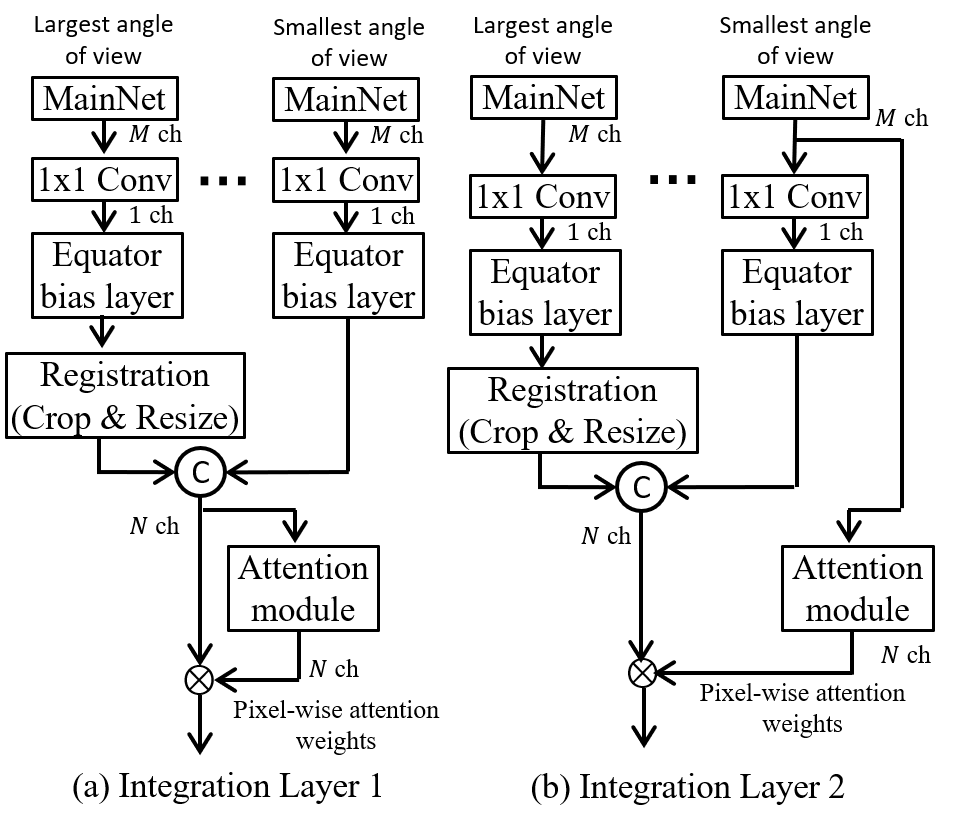}
    \end{center}
    \caption{Structures of integration layers. Details of “Attention module” is described in Table 1. $N$ corresponds to the number of angles of view.}
    \label{fig:integrationlayer}
\end{figure}

The structures of the integration layer for the 2D saliency maps estimated from the extracted patches with multiple angles of view at a viewing direction are shown in Fig. \ref{fig:integrationlayer}. Since the receptive fields are different among the angles of view, the 2D saliency maps with larger angles of view are cropped and resized to cover the same range with the smallest angle of view, using the relationship in Eq. \ref{eq:view_angle}. Then, they are integrated into a 2D saliency map using pixel-wise channel weighting, whose weights are calculated by an attention module. In the experiments, two integration structures were examined, as in Fig. \ref{fig:integrationlayer} (a) and (b). The attention weights are calculated from the concatenated saliency maps in (a), whereas the weights are obtained from image features in MainNet for the smallest angle of view in (b) since more information for importance of scales for each objects can be included in the image features. Since two structures of attention modules were also examined, the 4 types of architectures shown in Table \ref{table:arch_integrationlayer} were tested in the experiments.

\begin{table}[tb]
    \caption{Architectures of integration layer for multi-scale saliency estimation.}
    \label{table:arch_integrationlayer}
    \begin{center}
        \scalebox{0.75}{
        \begin{tabular}{c c c c c}
            \hline
            & Arch. 1 & Arch. 2 & Arch. 3 & Arch. 4 \\
            \hline
            Structure & Integration & Integration & Integration & Integration \\
            (Fig. \ref{fig:integrationlayer}) & Layer 1 & Layer 1 & Layer 2 & Layer 2 \\
            \hline
            Attention & 3x3 Conv ($C$ ch) & 1x1 Conv ($C$ ch) & 3x3 Conv ($C$ ch) & 1x1 Conv ($C$ ch) \\
            module & ReLU & ReLU & ReLU & ReLU \\
            & 2x2 Conv ($N$ ch) & 3x3 Conv ($C$ ch) & 2x2 Conv ($N$ ch) & 3x3 Conv ($C$ ch) \\
            & Softmax & ReLU & Softmax & ReLU \\
            & & 1x1 Conv ($4C$ ch) & & 1x1 Conv ($4C$ ch) \\
            & & ReLU & & ReLU \\
            & & 1x1 Conv ($N$ ch) & & 1x1 Conv ($N$ ch) \\
            & & Softmax & & Softmax \\
            \hline 
        \end{tabular}
        }
    \end{center}
\end{table}

\subsection{Normalization of Saliency Map}
\label{sec3:SalMapNormalization}

The output of the 2D saliency model is usually normalized by its norm, such as the L1 norm. If the 2D saliency model was used with the normalization to estimate the saliency for 2D patches extracted from ODI, the information of the dependence on the viewing direction for the extraction would be lost by the normalization at the integration to an ODI saliency map. Therefore, the 2D saliency model is used without the normalization at the output of the 2D saliency model. After the integration to the ODI saliency map, it is normalized by the L1 norm.

%\subsection{Metrics for ODI Saliency Map Estimation}
%\begin{figure}[tb]
%    \begin{center}
%        \includegraphics[width=\linewidth]{pics/ExampleOfSaliencyMapInSinusoidal.png}
%    \end{center}
%    \caption{Example of saliency map in ERP and in sinusoidal projection.}
%    \label{fig:9}
%\end{figure}

\section{Experimental Setup}

In the experiments, 'Head+Eye Saliency' in the Salient360 Dataset (ver.\ 2018) \cite{salient360_3} was used to evaluate ODI saliency maps. This dataset includes 85 ERP images with fixations obtained using HMD, whose field of view is 100$^\circ$ \cite{prosal}. Following \cite{salnet360}, the 25 images specified in the paper were used as test data, and the remaining 60 images were used as training and validation data. In the experiments for the comparison with conventional methods, only the 40 images in the training data (training data in salient360 ver.\ 2017) were used for fine-tuning of the network, which was same as \cite{salnet360}. The evaluation metrics used in the experiments were also same as the reference \cite{salnet360,salient360_toolbox}: the normalized scanpath saliency (NSS), the area under the curve (AUC), the correlation coefficient (CC), and the Kullback-Leibler divergence (KLD). Although these metrics have been also used in the evaluation of 2D saliency models, the metrics used in this paper were tailored to evaluate ODI saliency models by uniformly sampling on the sphere \cite{salient360_toolbox}, since the area at poles on the sphere is much smaller than in ERP\@. For these metrics, a higher value represents better performance except for KLD.

For the 2D saliency model in Fig. \ref{fig:2Dsalmodel}, DenseSal \cite{densesal} was adopted, which have achieved high performance in a saliency benchmark \cite{mit}. It was pre-trained with the ImageNet classification task, and was then fine-tuned by the 2D saliency-map estimation task with Salicon Dataset \cite{salicon} and OSIE Dataset \cite{osie}. In our experiments, this model was further fine-tuned with the extracted 2D images for multiple viewing directions with multiple angles of view from ODI training data in Salient360 Dataset. From the results of the preliminary experiments \cite{suzuki_SMC2018}, the interval of viewing directions was set to 45$^\circ$ in all the experiments in this paper. In this interval, 2D images were extracted at 26 viewing directions (8 horizontal directions $\times$ 3 vertical directions + 2 poles). At each of the viewing directions, multiple angles of view were used to extract the 2D images. The size of ODI in ERP was $800 \times 1600$ pixels, resized from the original sizes in the dataset, whereas the extracted 2D images were $500 \times 500$ pixels. The equator-bias layer in the saliency-map model was a scalar layer of $20 \times 20$ pixels with 5 channels corresponding to 5 elevation angles ($0^\circ, \pm45^\circ, \pm90^\circ$). The model was fine-tuned up to 5 epochs for the ODI dataset with the batch size of 1 due to the GPU memory limitation. Learning rates for MainNet and the bias layer were set to $10^{-5}$ and $10^{-4}$, respectively. KLD with L1 normalization and RMSProp were used for the loss function and the optimization method for training, respectively. The image features at the output of MainNet in Fig. \ref{fig:integrationlayer} was $M = 4416$ ch, and the intermediate channels in the attention modules was set to $C = 512$ ch in Table \ref{table:arch_integrationlayer}.

\section{Results}

\subsection{ODI Saliency Estimation with Learnable Equator Bias}

\begin{figure}[tb]
    \begin{center}
        \includegraphics[width=\linewidth]{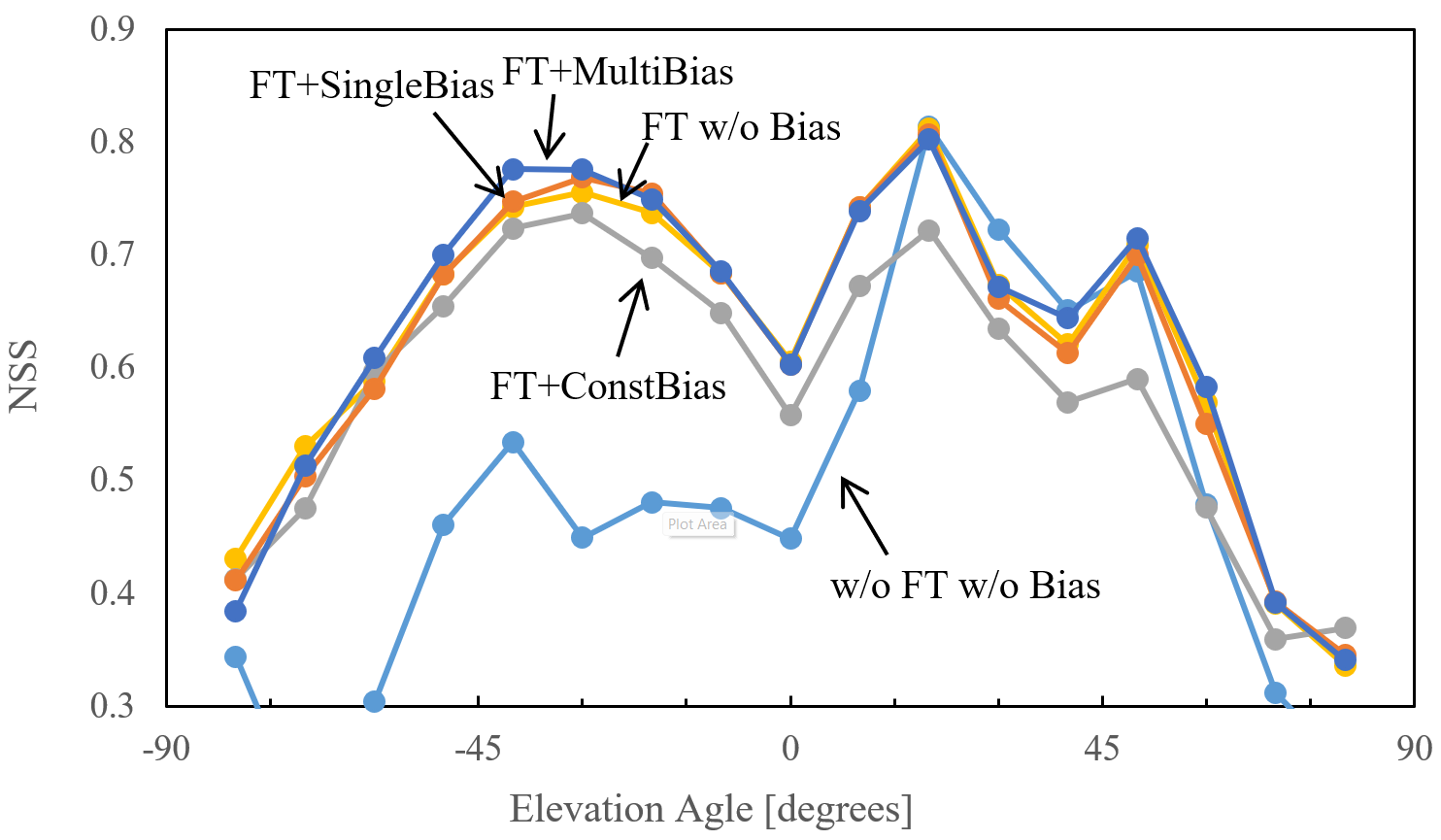}
    \end{center}
    \caption{Comparison at each elevation angle.}
    \label{fig:comp_ea}
\end{figure}

First, the effect of fine-tuning on the ODI dataset and the learnable equator-bias layer was examined in the condition of 2D-image extraction with the single angle of view 100$^\circ$. The results are shown in Table \ref{table:ft_and_equatorbias}. \textcolor{revised_color}{The experimental conditions were separated into two groups, MainNet without the fine-tuning to the ODI dataset (w/o FT in MainNet) and MainNet with the fine-tuning (w/ FT in MainNet).  For each condition in MainNet, several biases were compared.} The "Learned (single bias)" means that the 2D saliency model with a single-channel bias layer (same as the center-bias layer) was fine-tuned on the ODI dataset. The "Learned (multi-bias)" is the proposed method but using only a single angle of view for 2D-image extraction. The "Constant \textcolor{revised_color}{(Avg)}" bias means that the average equator bias over images in the ODI dataset was used for weighting the output of 2D saliency map. From the results, the fine-tuning \textcolor{revised_color}{of MainNet} was highly effective to improve the performance. The learnable bias layers were also effective \textcolor{revised_color}{in both the conditions in 'MainNet w/o FT' and 'MainNet w/ FT'}, especially in the multiple-channel equator bias which achieved the best performance \textcolor{revised_color}{in the condition of MainNet with the fine-tuning} in all the metrics. This means that leaning bias layer depending on the elevation angle for the 2D image extraction was useful to estimate the ODI saliency maps. To see this more clearly, NSS was calculated against the elevation angle of ODI in ERP, as shown in Fig. \ref{fig:comp_ea}. It can be seen from the figure that the fine-tuning with the multiple-channel bias was better than other conditions, especially around at $-45^\circ$.

\begin{table}[tb]
    \caption{Effect of fine-tuning on ODI dataset and equator-bias layer. FT: Fine-Tuning on ODI dataset.}
    \label{table:ft_and_equatorbias}
    \begin{center}
        \scalebox{0.9}{
        \begin{tabular}{c c c c c c}
            \hline
            \textcolor{revised_color}{MainNet} & Bias & NSS $\uparrow$ & AUC $\uparrow$ & CC $\uparrow$ & KLD $\downarrow$ \\
            \hline
            \textcolor{revised_color}{w/o FT} & \textcolor{revised_color}{w/o Bias} & 0.8711 & 0.7098 & 0.5634 & 0.7152 \\
            \textcolor{revised_color}{w/o FT} & \textcolor{revised_color}{Learned (single bias)} & \textcolor{revised_color}{0.8906} & \textcolor{revised_color}{0.7119} & \textcolor{revised_color}{0.5766} & \textcolor{revised_color}{0.7640} \\
            \textcolor{revised_color}{w/o FT} & \textcolor{revised_color}{Learned (multi-bias)} & \textcolor{revised_color}{0.8986} & \textcolor{revised_color}{0.7142} & \textcolor{revised_color}{0.5819} & \textcolor{revised_color}{0.7586} \\
            \textcolor{revised_color}{w/ FT} & Constant \textcolor{revised_color}{(Avg)} & 1.0097 & 0.7313 & 0.6601  & 0.7116 \\
            \textcolor{revised_color}{w/ FT} & \textcolor{revised_color}{w/o Bias} & 1.1083 & 0.7531 & 0.7295  & 0.3038 \\
            \textcolor{revised_color}{w/ FT} & Learned (single bias) & 1.1108 & 0.7527 & 0.7306 & 0.3021 \\
            \textcolor{revised_color}{w/ FT} & Learned (multi-bias) & $\mathbf{1.1601}$ & $\mathbf{0.7541}$ & $\mathbf{0.7345}$ & $\mathbf{0.3005}$ \\
            \hline
        \end{tabular}
        }

        % \begin{tabular}{c c c c c c}
        %     \hline
        %     FT & Bias & NSS $\uparrow$ & AUC $\uparrow$ & CC $\uparrow$ & KLD $\downarrow$ \\
        %     \hline
        %     No & No & 0.8711 & 0.7098 & 0.5634 & 0.7152 \\
        %     Yes & No & 1.1083 & 0.7531 & 0.7295  & 0.3038 \\
        %     Yes & Constant & 1.0097 & 0.7313 & 0.6601  & 0.7116 \\
        %     Yes & Learned (single bias) & 1.1108 & 0.7527 & 0.7306 & 0.3021 \\
        %     Yes & Learned (multi-bias) & $\mathbf{1.1601}$ & $\mathbf{0.7541}$ & $\mathbf{0.7345}$ & $\mathbf{0.3005}$ \\
        %     \hline
        % \end{tabular}
        
        %\includegraphics[width=\linewidth]{pics/EffectOfFineTuningAndEquatorBias.png}
    \end{center}
\end{table}

\subsection{Multi-scale Saliency Estimation}

Next, the effect of extracting 2D images in multiple angles of view (MAV) was studied for the integration layers in Table \ref{table:arch_integrationlayer}. The results are shown in Table \ref{table:multiscale}. The 4 architectures of the integration layer with MAV ($100^\circ, 110^\circ, 120^\circ$) were compared with the results with single angle of view (SAV) $100^\circ$. It can be seen from the table that the performance on all the architectures with MAV was better than that with SAV. Among the 4 architectures, "Arch.\ 4" achieved the best performance. This means that the image features from MainNet (DenseSal) with deeper attention modules were effective to calculate the attention weights on multiple scales for each pixel.

\begin{table}[tb]
    \caption{Effect of multi-scale method by extracting 2D images in multi-angles of view. SAV: Single Angle of View, MAV: Multi-Angles of View.}
    \label{table:multiscale}
    \begin{center}
        \begin{tabular}{l c c c c c}
            \hline
            \multicolumn{2}{c}{Methods} & NSS $\uparrow$ & AUC $\uparrow$ & CC $\uparrow$ & KLD $\downarrow$ \\
            \hline
            SAV & & 1.1601 & 0.7541 & 0.7345 & 0.3005 \\
            \hline
            \multirow{4}{*}{MAV} & Arch.\ 1 & 1.1638 & 0.7604 & 0.7789 & 0.2523 \\
            & Arch.\ 2 & 1.1728 & 0.7617 & 0.7879 & 0.2467 \\
            & Arch.\ 3 & 1.1732 & 0.7618 & 0.7881 & 0.2427 \\
            & Arch.\ 4 & $\mathbf{1.1937}$ & $\mathbf{0.7642}$ & $\mathbf{0.8010}$ & $\mathbf{0.2344}$ \\
            \hline
        \end{tabular}
    \end{center}
\end{table}

To examine the effective combination of angles of view, several combinations were tested, as shown in Table \ref{table:viewangles}. It can be seen from the results that smaller angles of view than the field of view in HMD ($100^\circ$) were not effective, as in ($80^\circ, 90^\circ, 100^\circ$). Since the performance with more than 3 angles of view decreased, the combination of ($100^\circ, 110^\circ, 120^\circ$) was the best on performance among the conditions tested. The reason for this might be that ($100^\circ, 110^\circ, 120^\circ$) was enough to capture the saliency for people, or that the larger angles of view detected the saliency which was not focused by people.

\begin{table}[tb]
    \caption{Comparison among angles of view for extraction.}
    \label{table:viewangles}
    \begin{center}
        \begin{tabular}{l c c c c c}
            \hline
            Angles of View & NSS $\uparrow$ & AUC $\uparrow$ & CC $\uparrow$ & KLD $\downarrow$ \\
            \hline
            (100, 110) & 1.1736 & 0.7617 & 0.7874 & 0.2478 \\
            (80, 90, 100) & 1.1438 & 0.7587 & 0.7678 & 0.2673 \\
            (90, 100,  110) & 1.1897 & 0.7612 & 0.7961 & 0.2386 \\
            (100, 110, 120) & $\mathbf{1.1937}$ & $\mathbf{0.7642}$ & $\mathbf{0.8010}$ & $\mathbf{0.2344}$ \\
            (100, 110, 120, 130) & 1.1540 & 0.7593 & 0.7758 & 0.2570 \\
            (100, 110, 120, 130, 140) & 1.1828 & 0.7631 & 0.7944 & 0.2394 \\
            \hline
        \end{tabular}
    \end{center}
\end{table}

\subsection{Comparison with Conventional Method}
To compare the performance with conventional methods, the proposed model was evaluated in the same experimental setup as the conventional methods following the description in \cite{salnet360} and \cite{salient360_toolbox}. \textcolor{revised_color}{In this experiment, two additional state-of-the-art (SOTA) 2D saliency models (DPNSal \cite{densesal} and DeepGaze IIE \cite{deepgaze2e}) in addition to DenseSal were tested as MainNet in the proposed method.} The results are shown in Table \ref{table:conventionalmethods}. The values for the conventional methods were taken from the reference \cite{salnet360}. As can be seen from the table, \textcolor{revised_color}{the proposed method with DenseSal achieved the best performance in all the metrics. Although DeepGaze IIE \cite{deepgaze2e} has shown the best performance in the 2D saliency benchmark \cite{mit_tuebingen}, the performance on the ODI saliency estimation was worse than DenseSal and DPNSal. Since DeepGaze IIE has frozen the base networks (ShapeNetC, EfficientNetB5, DenseNet201, and ResNext50) and has trained only a readout network, it would be difficult to be adapted to new datasets such as the ODI saliency dataset. The effectiveness of integrating multi-view angles (MAV) was confirmed for all the 2D saliency models tested.} The proposed method with DenseSal outperformed the conventional methods by a large margin.

\begin{table}[tb]
    \caption{Comparison with conventional methods.}
    \label{table:conventionalmethods}
    \begin{center}
        \begin{tabular}{l c c c c}
            \hline
            Methods & NSS $\uparrow$ & AUC $\uparrow$ & CC $\uparrow$ & KLD $\downarrow$ \\
            \hline
            %Mean Saliency Map & 0.3660 & 0.6390 & 0.5880 & 0.4410 \\
            TU Munich \cite{TUMunich} & 0.805 & 0.726 & 0.579 & 0.449 \\
            SJTU \cite{sjtu} & 0.918 & 0.735 & 0.532 & 0.481 \\
            CDSR \cite{cdsr} & 0.936 & 0.736 & 0.538 & 0.508 \\
            GBVS360-Eq \cite{prosal} & 0.851 & 0.714 & 0.527 & 0.698 \\
            SalNet360 \cite{salnet360} & 0.757 & 0.702 & 0.536 & 0.487 \\
            \hline
            \multicolumn{5}{l}{\textcolor{revised_color}{Proposed (MainNet: DenseSal \cite{densesal})}} \\
            \multicolumn{1}{c}{SAV} & 1.1405 & 0.7537 & 0.7622 & 0.2781 \\
            \multicolumn{1}{c}{MAV} & $\mathbf{1.1751}$ & $\mathbf{0.7623}$ & $\mathbf{0.7891}$ & $\mathbf{0.2464}$ \\
            %Proposed (SAV) & 1.1601 & 0.7541 & 0.7345 & 0.3005 \\
            %Proposed (MAV) & $\mathbf{1.1937}$ & $\mathbf{0.7642}$ & $\mathbf{0.8010}$ & $\mathbf{0.2344}$ \\
            \multicolumn{5}{l}{\textcolor{revised_color}{Proposed (MainNet: DPNSal \cite{densesal})}} \\
            \multicolumn{1}{c}{\textcolor{revised_color}{SAV}} & \textcolor{revised_color}{1.0858} & \textcolor{revised_color}{0.7509} & \textcolor{revised_color}{0.7178} & \textcolor{revised_color}{0.3112} \\
            \multicolumn{1}{c}{\textcolor{revised_color}{MAV}} & \textcolor{revised_color}{1.1105} & \textcolor{revised_color}{0.7539} & \textcolor{revised_color}{0.7296} & \textcolor{revised_color}{0.3061} \\
            \multicolumn{5}{l}{\textcolor{revised_color}{Proposed (MainNet: DeepGazeIIE \cite{deepgaze2e})}} \\
            \multicolumn{1}{c}{\textcolor{revised_color}{SAV}} & \textcolor{revised_color}{0.9446} & \textcolor{revised_color}{0.7275} & \textcolor{revised_color}{0.6169} & \textcolor{revised_color}{0.4053} \\
            \multicolumn{1}{c}{\textcolor{revised_color}{MAV}} & \textcolor{revised_color}{0.9836} & \textcolor{revised_color}{0.7359} & \textcolor{revised_color}{0.6465} & \textcolor{revised_color}{0.3770} \\
            \hline
        \end{tabular}
        
        % \begin{tabular}{l c c c c}
        %     \hline
        %     Methods & NSS $\uparrow$ & AUC $\uparrow$ & CC $\uparrow$ & KLD $\downarrow$ \\
        %     \hline
        %     %Mean Saliency Map & 0.3660 & 0.6390 & 0.5880 & 0.4410 \\
        %     TU Munich \cite{TUMunich} & 0.805 & 0.726 & 0.579 & 0.449 \\
        %     SJTU \cite{sjtu} & 0.918 & 0.735 & 0.532 & 0.481 \\
        %     CDSR \cite{cdsr} & 0.936 & 0.736 & 0.538 & 0.508 \\
        %     GBVS360-Eq \cite{prosal} & 0.851 & 0.714 & 0.527 & 0.698 \\
        %     SalNet360 \cite{salnet360} & 0.757 & 0.702 & 0.536 & 0.487 \\ 
        %     Proposed (SAV) & 1.1405 & 0.7537 & 0.7622 & 0.2781 \\
        %     Proposed (MAV) & $\mathbf{1.1751}$ & $\mathbf{0.7623}$ & $\mathbf{0.7891}$ & $\mathbf{0.2464}$ \\
        %     %Proposed (SAV) & 1.1601 & 0.7541 & 0.7345 & 0.3005 \\
        %     %Proposed (MAV) & $\mathbf{1.1937}$ & $\mathbf{0.7642}$ & $\mathbf{0.8010}$ & $\mathbf{0.2344}$ \\
        %     \hline
        % \end{tabular}

        %\includegraphics[width=\linewidth]{pics/ComparisonWithConventionalMethod.png}
    \end{center}
\end{table}

\subsection{Examples of Saliency Maps}

\begin{figure}[tb]
    \begin{center}
        \includegraphics[width=\linewidth]{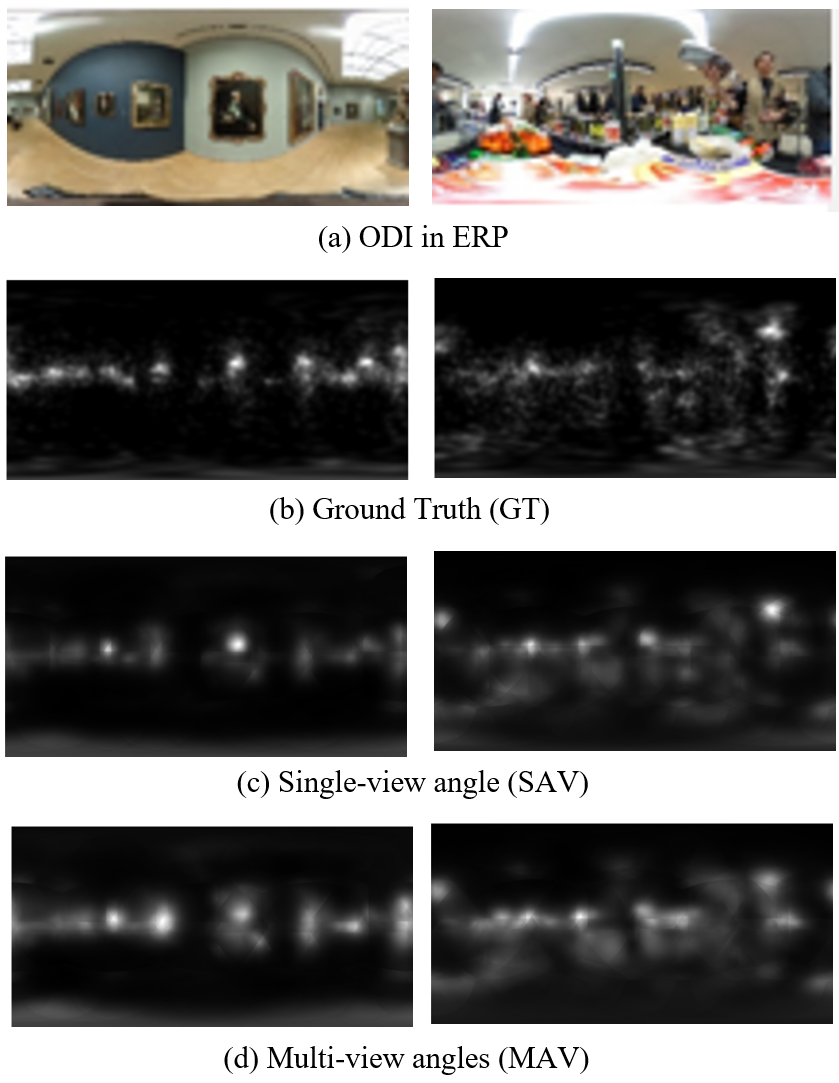}
    \end{center}
    \caption{Sample saliency maps for DOI in ERP\@.}
    \label{fig:sample_ODISalMap}
\end{figure}

\begin{figure}[tb]
    \begin{center}
        \includegraphics[width=\linewidth]{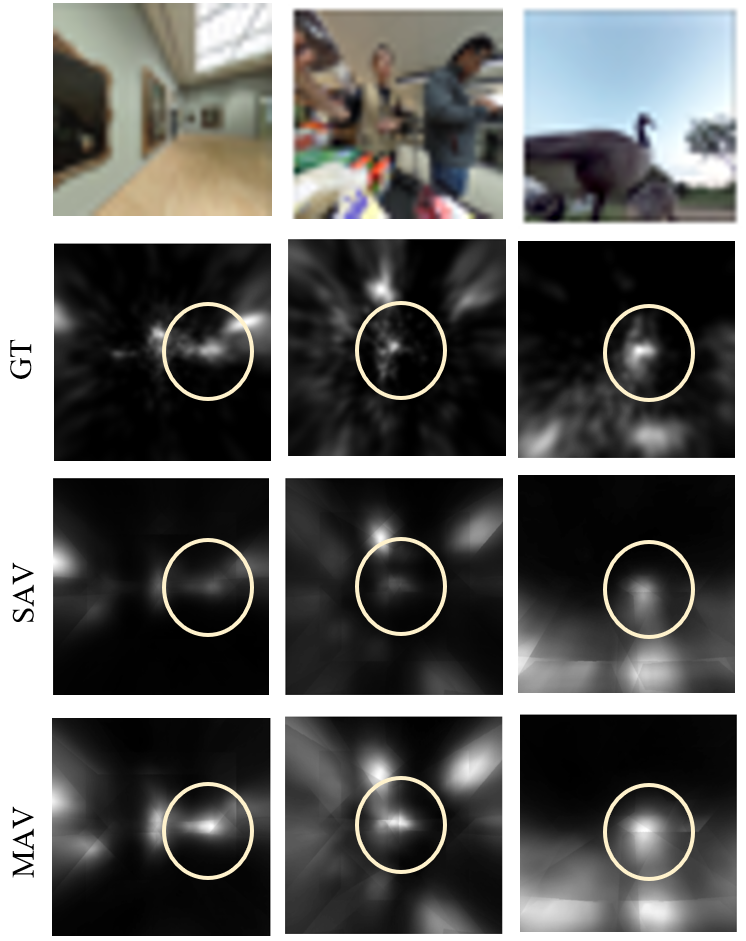}
    \end{center}
    \caption{2D saliency maps extracted from ODI saliency maps.}
    \label{fig:sample_2DSalMap}
\end{figure}

The examples of saliency maps estimated by the proposed methods (SAV and MAV with Arch.\ 4 in Table \ref{table:multiscale}) are shown in Fig. \ref{fig:sample_ODISalMap}. As can be seen from the sample images, most of the salient regions were detected both in SAV and MAV. To see more details, 2D images were extracted from the estimated ODI saliency maps, as shown in Fig. \ref{fig:sample_2DSalMap}. In the left column of the figure, a small painting was fixated in the ground truth (GT), whereas SAV estimated lower saliency than GT. By using the multi-scale method with MAV, high saliency was estimated to the small painting. Similarly, a left hand of a human in the middle column example and a small head of a bird in the right column example were provided higher saliency by MAV than SAV as in GT.

\section{Conclusions}

In this paper, a novel ODI saliency estimation model was proposed, based on the well-established 2D saliency model pretrained on large 2D saliency databases. By extracting overlapping 2D patches from ODI in various viewing directions with multiple angles of view, the saliency of objects was accurately estimated in the optimal scales. To compensate the difference between the prior distributions for 2D saliency maps (center bias) and ODI saliency maps (equator bias), the 2D saliency model was pretrained with a center-bias layer to explicitly represent the prior distribution and was fine-tuned to an ODI saliency dataset by replacing the center-bias layer to an equator-bias layer composed of multiple channels corresponding to elevation angles for 2D-patch extraction. From the experiments, it was confirmed that the accuracy of the ODI saliency map estimation was improved by the proposed method by a large margin.

\textcolor{revised_color}{
    One of the limitations of the ODI saliency prediction is that the database for this task has been relying on a single small dataset only including 60 images for training and 25 images for testing, since the publicly available database for this task is currently limited to this database. Thus, the presented results are limited to the evaluation for only the 25 omni-directional images. Therefore, the more evaluation would be desired in the future research by creating larger databases for this task. Furthermore, since the proposed method is relying on the extraction of multiple 2D images from an omni-directional image, the computational cost for the saliency prediction was higher than the methods which predict saliency directly from omni-directional images. Therefore, the suppression of the computational cost would be another future research direction.
}

%\bibliographystyle{ieicetr}% bib style
%\bibliography{}% your bib database

\profile{Takao Yamanaka}{
%\profile*{Takao Yamanaka}{
        received the B.S., M.S., and Ph.D. degrees in Electrical Engineering from Tokyo Institute of Technology in 1996, 1998, and 2004, respectively. During 1998-2000, he worked in Canon Inc. After working in Texas A\&M University as a post-doctoral fellow, he joined Sophia University in 2006, where he is currently an associate professor in Department of Information and Communication Sciences.
}% without picture of author's face

\profile{Tatsuya Suzuki}{
%\profile*{Tatsuya Suzuki}{
    received the B.S. and M.S. degrees in Computer Science from Sophia University in 2018 and 2020, respectively. During his master's program, he worked on saliency estimation for omni-directional images.
}% without picture of author's face

\profile{Taiki Nobutsune}{
    received the B.S degree in Computer Science from Sophia University in 2023. He worked on multi-scale saliency estimation for omni-directional images in his undergraduate project.
}% without picture of author's face

\profile{Chenjunlin Wu}{
    received the M.S. degree in Computer Science from Sophia University in 2022. During his master's program, he worked on saliency estimation for both 2D images and omni-directional images.
}% without picture of author's face

\end{document}